\pdfoutput=1

\documentclass[11pt]{article}

\usepackage[]{acl}

\usepackage{times}
\usepackage{latexsym}

\usepackage[T1]{fontenc}

\usepackage[utf8]{inputenc}

\usepackage{microtype}
\usepackage{tabularx}
\usepackage{booktabs}
\usepackage{boldline}
\usepackage{subcaption}
\usepackage{multirow}
\usepackage{float}
\usepackage{xcolor}
\usepackage{graphicx}
\usepackage{dialogue}
\usepackage{csquotes}
\usepackage{listings}
\usepackage[para]{threeparttable}
\usepackage{soul} 
\usepackage{enumitem}
\usepackage{mathtools}
\usepackage{adjustbox}
\usepackage{algorithm}
\usepackage[noend]{algpseudocode}
\usepackage{makecell}
\usepackage{amsfonts}
\usepackage{cleveref}
\usepackage{varwidth}
\usepackage{etoolbox}
\usepackage{amssymb}
\usepackage{pifont}
\newcommand{\cmark}{\ding{51}}
\newcommand{\xmark}{\ding{55}}

\algrenewcommand\algorithmicrequire{\textbf{Input:}}
\algrenewcommand\algorithmicensure{\textbf{Output:}}

\usepackage{colortbl}
\usepackage{color}

\usepackage{CJKutf8} 

\newcommand\blfootnote[1]{%
  \begingroup
  \renewcommand\thefootnote{}\footnote{#1}%
  \addtocounter{footnote}{-1}%
  \endgroup
}

\definecolor{grayout}{gray}{0.9}

\title{Aligning Large Language Models by On-Policy Self-Judgment}

\author{
    \textbf{Sangkyu Lee}$^{1,\ast}$ \quad \textbf{Sungdong Kim}$^{2,3,\dagger}$ \quad \textbf{Ashkan Yousefpour}$^{1}$ \\
    \textbf{Minjoon Seo}$^{3}$ \quad 
    \textbf{Kang Min Yoo}$^{2,4}$ \quad \textbf{Youngjae Yu}$^{1,\dagger}$ \\
    Yonsei University$^{1}$ \quad NAVER Cloud$^{2}$ \quad KAIST AI$^{3}$ \quad SNU AI Center$^{4}$ \\
}

\begin{document}

\def\method{\textsc{Self-Judge}}
\maketitle

\blfootnote{\textsuperscript{$\ast$} Work done during internship at NAVER Cloud} 
\blfootnote{\textsuperscript{$\dagger$} Corresponding authors}

\begin{abstract}

Existing approaches for aligning large language models with human preferences face a trade-off that requires a separate reward model (RM) for on-policy learning. In this paper, we present a novel alignment framework, \method{} that (1) does on-policy learning and 2) is parameter efficient, as it does not require an additional RM for evaluating the samples for on-policy learning. To this end, we propose Judge-augmented Supervised Fine-Tuning (JSFT) to train a single model to act as both a policy and a judge. Specifically, we view the pairwise judgment task, choosing the better response from a response pair, as a special case of the instruction-following task. The resulting model can judge preferences of on-the-fly responses from current policy initialized from itself. Experimental results show the efficacy of \method{}, outperforming baselines in preference benchmarks. We also show that the rejecting sampling by itself can improve performance further without an additional evaluator\footnote{The code is available at \href{https://github.com/oddqueue/self-judge}{github.com/oddqueue/self-judge}}.
\end{abstract}

\section{Introduction}

Research on aligning Large Language Models (LLMs) with human preference has increasingly gained attention in recent years~\cite{askell2021general, ouyang2022training, bai2022training, rafailov2023direct}. Reinforcement Learning from Human Feedback (RLHF) is the most dominant approach for the alignment of LLMs for human preferences~\cite{ziegler2020finetuning}. It utilizes a reward model (RM) to estimate the human preference scores for the generated responses from policy. The policy is updated with \textit{on-policy} Reinforcement Learning (RL) to maximize the estimated rewards of sampled responses regularized with the KL divergence between the current policy and reference policy. However, RLHF requires a complex setup for on-policy updates because of its simultaneous use of an RM with the reference model.

\begin{figure}[t]
  \centering
   \includegraphics[width=\columnwidth]{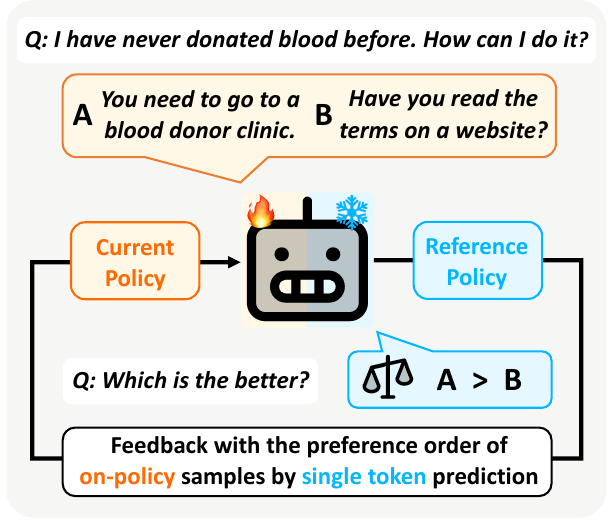}
  \caption{In our framework, \method{}, a single model is trained not only to generate responses but also to perform a judgment task, where it selects the better of the two responses through a single token prediction. This enables on-policy self-training by performing judgments on current policy for improving itself.}
  \label{fig:overview}
\end{figure}
\begin{figure*}[t]
  \centering
  \includegraphics[width=\textwidth]{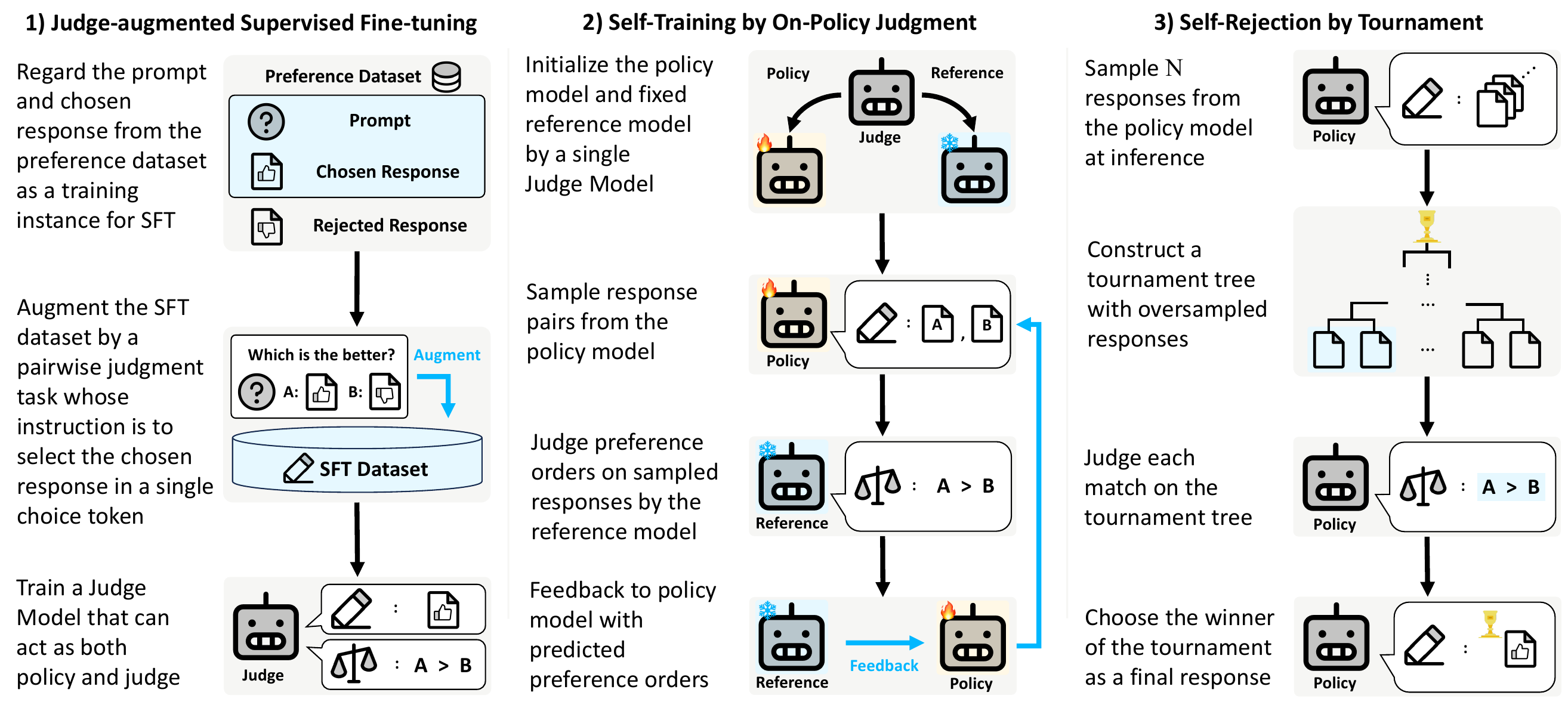}
  \caption{An overview of \method{}. 1) We train an LLM to act as a Judge Model (JM), which can both generate responses and compare response pairs. We train the JM with a SFT dataset augmented with the pairwise judgment task where the better response can be selected by a single token. 2) We initialize a policy and a fixed reference model from the trained JM. Then, the policy model samples response pairs, and the reference model performs judgments on the pairs for giving feedback with preference orders. 3) We perform a rejection sampling by a tournament on responses from the policy through the judgments by itself for further improvements at inference time.}
  \label{fig:architecture}
\end{figure*}

In contrast, a line of research~\cite{rafailov2023direct, azar2023general} proposes discarding on-policy learning and optimizing from preference orders in static datasets without RMs (i.e., \textit{offline} learning), whereas other studies propose constructing datasets with responses sampled from the initial policy and labeled by separated evaluators (i.e., \textit{off-policy} learning)~\cite{zhao2023slic, liu2023statistical, xu2023some, gulcehre2023reinforced, dong2023raft}. However, offline learning can lead to sub-optimal results due to the lack of exploration~\cite{mediratta2023generalization}, and off-policy learning could cause degeneration if an appropriate replay buffer strategy is not used~\cite{zhang2017deeper}.

In this paper, we present \textbf{\method{}}, a simplified \textit{on-policy} training scheme that trains a \textit{single} model to perform on-the-fly feedback for current policy for improving \textit{itself}. To this end, we introduce Judge-augmented Supervised Fine-tuning (JSFT) to obtain a model that can both generate a response and perform pairwise comparison, as shown in Figure~\ref{fig:overview}. Specifically, we regard the pairwise judgment task, choosing the better response from a response pair, as a special case of the instruction-following task, which can be answered in a \textit{single token} and optionally with a rationale. \method{} initializes the current policy and reference policy from JSFT's resulting model. \method{} samples a response pair from the current policy and chooses a better response in the pair by the reference policy, then updates the current policy with preference orders without an RM~\cite{rafailov2023direct, zhao2023slic, azar2023general}.

Experimental results show that \method{} outperforms RLHF and other offline and off-policy approaches~\cite{rafailov2023direct, liu2023statistical, gulcehre2023reinforced, dong2023raft} on preference benchmarks. Unlike existing approaches, \method{} leverages on-policy learning while not introducing an additional evaluator. The results demonstrate the effectiveness and parameter efficiency of \method{} performing on-policy \textit{self-training} for LLM alignment. Furthermore, we show that \method{} can maximize performance through \textit{self-rejection} that selects the best response from its own responses using its judgment capabilities learned through JSFT. In particular, the performance gains are significant when the pairwise judgment task of JSFT involves comparisons based on principles with rationale for the decision.

In summary, our main contributions are:

\begin{itemize}
    \setlength\itemsep{0em}
  \item We propose a parameter-efficient on-policy learning framework, \method{}, with JSFT for obtaining an initial policy that can judge. 
  \item We analyze the efficacy of the JSFT for judgment and suggest the best practices to exploit the improved judgment ability.
  \item We show resulting models from JSFT can self-improve by acting as a judge: on-policy self-training and self-rejection at inference time.
\end{itemize}

\section{Preliminaries}

\paragraph{Aligning by Reinforcement Learning}
To align LLMs with human preferences, RLHF~\cite{ziegler2020finetuning} follows three stages: 1) obtaining an initial policy $\pi_{\text{ref}}$ through SFT from the pre-trained LLM, 2) training an RM from human preference triplets $(x, y_w, y_l)$, where $y_w$ is a chosen response, and $y_l$ is a rejected response for a given prompt $x$, 3) fine-tuning the initial policy by Proximal Policy Optimization (PPO)~\cite{schulman2017proximal}, with KL divergence between current policy $\pi$ and reference policy $\pi_{\text{ref}}$ as a regularization for the reward maximization objective. Generally, RMs are trained by Bradley-Terry model~\cite{bradley1952rank}, which minimizes negative log-likelihood of score difference, $\text{log}\sigma(r_{\theta}(x, y_w) - r_{\theta}(x, y_l))$, to compute a pointwise scalar reward for the response sampled from the current policy. This \textit{on-policy} rollout procedure provides the frequent \textit{exploration} of responses but introduces an additional training stage and memory usage due to the RMs.

\paragraph{Aligning from Preference Orders}
It has been observed that RM is susceptible to language biases such as the response length, preferring longer responses over shorter ones~\cite{shen2023loose}. \citet{askell2021general} suggest a language modeling loss on preferred context $(x, y_w) $ during the training of the RM to address this issue. This approach, however, necessitates another pre-training stage of the RM. Direct Preference Optimization (DPO)~\cite{rafailov2023direct} and Sequence Likelihood Calibration with Human Feedback (SLiC-HF)~\cite{zhao2023slic} introduce objectives that can be trained by only preference orders $y_w \succ y_l$ without scalar rewards and need for RMs. However, \textit{offline} learning methods optimized on static datasets inherently reach sub-optimal results compared to \textit{online} learning~\cite{mediratta2023generalization}. \citet{xu2023some, yuan2024self} propose an online learning approach that iteratively constructs datasets by responses sampled from the policy. However, this \textit{off-policy} approach with a large buffer size can induce performance degeneration~\cite{zhang2017deeper}.
\section{\method{}}
In this section, we describe our proposed alignment framework, \textbf{\method{}}, which utilizes a \textit{single} model that acts as both policy and judge: sampling responses and judgment over response pairs. It can provide feedback on the response pairs sampled from the current policy for improving itself in an \textit{on-policy} manner and also perform rejection sampling by itself, as depicted in \Cref{fig:architecture}.

\subsection{Judge Model}
\begin{figure}[t]
  \centering
  \includegraphics[width=\columnwidth]{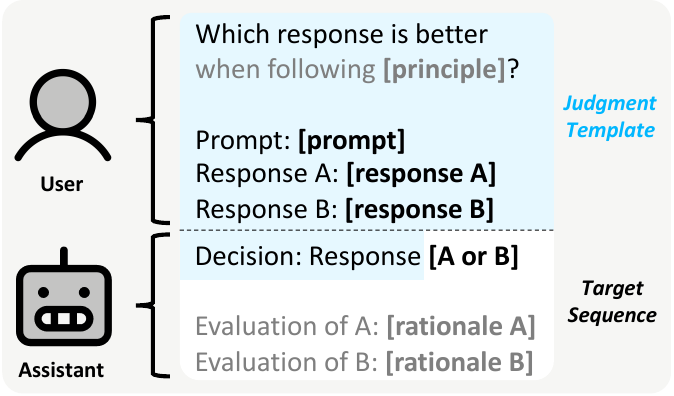}
  \caption{An example of a judgment template $\mathcal{C}$. The judgment template asks which of the two responses is better for a given prompt and requests to select the judge token $\mathcal{J} \in \{\mathcal{A}, \mathcal{B}\}$ corresponding to the better response. Optionally, a principle for the judgment can be added to the judgment template, and the rationale can be included in the target sequence for training.}
  \label{fig:judgment_example}
\end{figure}

Using LLMs to evaluate responses of another LLM, \textit{LLM-as-a-judge}, is shown promising~\cite{zheng2023judging, bai2022constitutional, kim2023prometheus}. Inspired by these studies, we leverage the generative pairwise evaluator, which we refer to as \textbf{Judge Model (JM)}, for aligning LLMs with human preference~\cite{pmlr-v162-ethayarajh22a, zhao2023slic, liu2023statistical}. Unlike an RM, which produces a scalar score for a single response, JM simply chooses the better response between the two responses. Specifically, the JM $\pi$ is trained by maximizing the log-likelihood of the judge token $\mathcal{J} \in \{\mathcal{A}, \mathcal{B}\}$ corresponding ground-truth preference label for a given judgment template $\mathcal{C}$. \Cref{fig:judgment_example} illustrates an example of a judgment template.

\paragraph{Judge-augmented Supervised Fine-tuning}
Typically, the chosen response $y_w$ is used for target sequence of prompt $x$ in training preference dataset $\mathcal{D}$ on the SFT stage, i.e., $\mathcal{D}_{\text{SFT}} = \{(x, y_w) | (x, y_w, y_l) \in \mathcal{D}\}$. In addition, we treat the pairwise judgment task as a special case of instruction-tuning, e.g., $\mathcal{D}_{\text{Judge}} = \{\big(\mathcal{C}(x, y_w, y_l), \mathcal{J}\big) | (x, y_w, y_l) \in \mathcal{D}\}$. As a result, we train the pairwise judgment task as a response generation using the augmented dataset $\mathcal{D}^+_{\text{SFT}} = \mathcal{D}_{\text{SFT}} \cup \mathcal{D}_{\text{Judge}}$. We refer to the process of fine-tuning the LLMs on the augmented dataset $\mathcal{D}^+_{\text{SFT}}$ as \textbf{Judge-augmented Supervised Fine-tuning (JSFT)}. With JSFT, we can obtain a JM that can not only compare pairwise preferences but also generate responses. Also, we can expect a better understanding of language-relevant features on preference comparison since the JM is trained to mimic \textit{good} behavior $y_w$, similar to the observation found on RMs~\cite{askell2021general}. 

\paragraph{Principle-aware Judgment with Rationale}
One advantage of JM compared to the conventional method with RM is that the JM can adapt the judgment template $\mathcal{C}$ according to multiple aspects of human preferences, reflecting diverse principles~\cite{sun2023salmon, cui2023ultrafeedback}. Also, the generative nature of JM makes it easy to utilize rationale, the justification for the judgment in natural language, for more precise judgments, similar to the observations on instruction-tuning of LLMs~\cite{mukherjee2023orca}. For a given principle $p \in \mathcal{P}$ where the $\mathcal{P}$ is a principle set, we devise different judgment templates $\mathcal{C}_p$ for each $p$. Also, we can optionally include a rationale $\mathcal{R}$ in the target sequence for the judgment task when it is available. The JM first produces a judge token $\mathcal{J} \in \{\mathcal{A}, \mathcal{B}\}$ which corresponds to the ground-truth label of the better response, then rationale $\mathcal{R}$ is followed. Formally, $\pi(\mathcal{J}, \mathcal{R} \mid \mathcal{C}_p(x, y_w, y_l))$. As a result, principle-aware judgment can be conducted by simply adjusting the judgment template. 

\paragraph{Judging Self-Generated Responses}
Since JM is trained by JSFT, it can judge its own response by playing the roles of the policy and judge: this is the intuition behind the name \textbf{\method}. As a policy, the JM first samples responses $y_a, y_b$ for a given $x$. Then, the JM, as a judge, chooses a better response between the two responses, $\pi(\mathcal{J} \mid \mathcal{C}(x, y_a, y_b))$. More precisely, it averages the likelihoods of corresponding judge tokens, $\mathcal{J}$, utilizing a position-swapped judgment template to mitigate position bias~\cite{vicuna2023}. From the relative likelihoods of these judge tokens for each response, a pseudo-preference triplet label $(x, \hat{y}_w, \hat{y}_l)$ for the \textit{self-training} and \textit{self-rejection} can be deduced. For the principle-aware judgments, we can first check the winning rate across principles and then the mean likelihood of judge tokens across principles for deciding orders when a tie happens.

\subsection{Self-Training by On-Policy Judgment}
\citet{pmlr-v162-ethayarajh22a} uses JM's predicted likelihood of judge tokens as the reward for on-policy evaluation in RLHF framework, comparing with a \textit{blank} baseline response~\footnote{\href{https://huggingface.co/stanfordnlp/SteamSHP-flan-t5-xl}{huggingface.co/stanfordnlp/SteamSHP-flan-t5-xl}}.
This approach has a significant limitation; the likelihood inferred by a language model cannot be regarded as confidence without calibration~\cite{zhou2023navigating, zhu2023calibration}. Conversely, to leverage JMs properly, we adopt objectives that perform optimizations on preference orders since these objectives do not require pointwise scalar scores~\cite{rafailov2023direct, zhao2023slic, azar2023general}. In our framework, we regard the reference policy $\pi_\text{ref}$ as a judge to evaluate the preference order between generated samples from current policy $\pi_{\theta}$, as follows:
\begin{equation*}
\begin{split}
&y_a \sim \pi_{\theta}(\cdot \mid x), y_b \sim \pi_{\theta}(\cdot \mid x),\\
&(x, \hat{y}_w, \hat{y}_l) \leftarrow \pi_\text{ref}(\mathcal{J} \mid \mathcal{C}(x, y_a, y_b)). \\
\end{split}
\end{equation*}

That is, a single JM with JSFT is initialized for both $\pi_{\theta}$ and $\pi_\text{ref}$, but the latter one is frozen for the likelihood normalization~\cite{liu2023statistical} and on-policy judgments. This setup enjoys \textit{on-policy} learning without introducing an additional model for the evaluation as RLHF. If we adopt the DPO objective, the following loss is used for optimization, where $\sigma$ is the sigmoid function, and $\beta$ is a coefficient for KL divergence regularization,
\begin{equation*}
\begin{split}
- \mathbb{E} \big[\log \sigma \big(\beta \log \cfrac{\pi_{\theta}(\hat{y}_w|x)}{\pi_{\text{ref}}(\hat{y}_w|x)} - \beta \log \cfrac{\pi_{\theta}(\hat{y}_l|x)}{\pi_{\text{ref}}(\hat{y}_l|x)}\big)\big]. \\
\end{split}
\end{equation*}

\subsection{Self-Rejection by Tournament}
RMs can also be utilized at inference time for rejection sampling such as \textit{Best-of-$N$} sampling~\cite{stiennon2020learning}, selecting the best response according to the reward score among oversampled $N$ responses. However, this approach requires a separate RM in addition to the policy model during inference. In contrast, JM trained within \method{} does not require an additional RM for evaluating responses, as it can evaluate the self-generated responses by itself. 

However, judging all the possible comparison pairs to get the average winning rate requires $\mathcal{O}(N^2)$ forward passes for ${N \choose 2}$ comparisons. Thus, we adopt the \textit{tournament}~\cite{zhao2023slic, liu2023statistical} for rejection sampling at inference time. We construct a tournament tree whose leaf nodes are sampled responses, and the non-leaf nodes are chosen by the winner on judgment between the child nodes. Since the tournament tree has less than $N-1$ non-leaf nodes, we can find the best response by $\mathcal{O}(N)$ forward passes, the same as the Best-of-$N$ sampling with a separated RM.
\section{Experimental Setup}

\begin{table*}[t]
    \centering
    \adjustbox{max width=\textwidth}{%
    \begin{tabular}{l|cc|cc|ccc}
        \toprule
        \textbf{Method} & \textbf{Policy} & \textbf{Evaluator} & \makecell{\textbf{On-Policy}\\ \textbf{Learning}} & \makecell{\textbf{On-Memory}\\ \textbf{Parameters}} & \makecell{\textbf{AlpacaEval}\\ \small{(\% Win)}} & \makecell{\textbf{VicunaEval}\\ \small{(\% Win)}} & \makecell{\textbf{MT-Bench} \\\small{(Score)}} \\
        \midrule
        SFT & \xmark & \xmark & \xmark & $p$ & 24.75 & 50.00 & 4.63 \\
        \midrule
        DPO & SFT & \xmark & \xmark & $2p$ &\underline{35.14} & 60.63 & \underline{4.73} \\
        RSO & SFT & JM & \xmark & $2p$ & 34.27 & \underline{64.38} & 
        4.42\\
        ReST & SFT & RM & \xmark & $p$ & 27.43 & 55.00 & 
        4.53\\
        RAFT & SFT & RM & \xmark & $p$ & 32.50 & 59.38 & 
        4.43\\
        RLHF & SFT & RM & \cmark & $3p$ & 33.46 & 53.75 & 
        4.29\\
        \midrule
        \makecell{\textbf{\method{}} \\ \small{(Ours)}} & \multicolumn{2}{c|}{JM} & \cmark & $2p$ & \textbf{44.88} & \textbf{76.25} & \textbf{4.80} \\
        \bottomrule
    \end{tabular}}
    \caption{Evaluation results of models trained on HH-Helpful~\cite{bai2022training}. The best result and second best result on each benchmark are represented as bold and underline. We report theoretical memory usage of model parameters required for each method where $p$ denotes the number of parameters of the base model. We use \texttt{base} and \texttt{online} splits but used the \texttt{online} split only for constructing the training instances of SFT. \method{} outperforms baselines on all benchmarks with a single JM, which can act as both policy and judge.}
    \label{table:hh_helpful_main}
\end{table*}

In this section, we discuss the experimental setup for validating our proposed framework, \method{}. We use two datasets: Anthropic-HH~\cite{yuan2023rrhf} and UltraFeedback~\cite{cui2023ultrafeedback}. In the experiments, we choose DPO~\cite{rafailov2023direct}, RSO~\cite{liu2023statistical}, ReST~\cite{gulcehre2023reinforced}, RAFT~\cite{dong2023raft}, and
RLHF~\cite{ziegler2020finetuning} as baselines for comparing with \method{}. We evaluate the resulting models from each experiment by AlpacaEval~\cite{alpaca_eval}, VicunaEval~\cite{vicuna2023}, and MT-Bench~\cite{zheng2023judging}. The implementation details are in \Cref{sec:implementation_details}.

\subsection{Datasets}

\paragraph{Anthropic-HH}
Anthropic-HH~\cite{bai2022training} is a human preference dataset on 170k dialogues, which consists of two subsets, HH-Helpful and HH-Harmless, which are labeled by the helpfulness and harmlessness principle. We focus on the HH-Helpful to better isolate and understand the benefits of \method{} due to the conflicting nature of helpfulness and harmlessness principles~\cite{bai2022training}. HH-Helpful contains various data splits corresponding to the development stages of an AI assistant. We use the \texttt{base} split and include the pair $(x, y_w)$ from the \texttt{online} split in the SFT dataset for analyzing transition effects on JSFT.

\paragraph{UltraFeedback}
As obtaining human-labeled feedback is costly, utilizing AI feedback is a widely investigated alternative~\cite{bai2022constitutional, sun2023salmon, openai2023gpt4}. UltraFeedback~\cite{cui2023ultrafeedback} is one of the datasets that consists of AI feedback where GPT-4~\cite{openai2023gpt4} rates responses obtained from four different language models for the 64k prompts, based on four principles of helpfulness, instruction-following, truthfulness, and honesty. Each rating contains the rationale obtained from GPT-4, which represents an explanation for the quality and rating of the corresponding response based on the given principle. We randomly split 10\% of the prompts of the dataset and use them as a test set for further analysis.

\subsection{Baselines}

We choose \textbf{DPO}~\cite{rafailov2023direct} as an offline learning baseline and use DPO objective for self-training in \method{}. We include \textbf{RSO}~\cite{liu2023statistical} with the DPO objective as an off-policy learning baseline, which utilizes the JM as a separate evaluator. We include baselines that utilize RMs, \textbf{ReST}~\cite{gulcehre2023reinforced} and \textbf{RAFT}~\cite{dong2023raft} for off-policy learning approach and \textbf{RLHF}~\cite{ziegler2020finetuning} for on-policy learning approach. We choose Llama-2-7B~\cite{touvron2023llama2} as a base model for all experiments for fair comparisons.

\subsection{Evaluations}

We evaluate the resulting models based on the three benchmarks, AlpacaEval~\cite{alpaca_eval}, VicunaEval~\cite{vicuna2023}, and MT-Bench~\cite{zheng2023judging}. \textbf{AlpacaEval} is an alignment benchmark that compares the quality of two responses on 805 questions sampled from a diverse dataset, rated by GPT-4~\cite{openai2023gpt4}. We use the \texttt{text-davinci-003}~\cite{ouyang2022training} as the baseline model for measuring winning rates. \textbf{VicunaEval} is another benchmark utilizing GPT-4 as a judge for comparing two models' responses from 80 questions on various topics. We use the SFT model on each dataset as the baseline model for measuring winning rates. \textbf{MT-Bench} is a multi-turn benchmark consisting of 80 instances of 2-turn questions from 8 different domains, which is evaluated by GPT-4 on a scale of 1 from 10. We report the average scores obtained on each turn.
\section{Experimental Results}

\subsection{Main Results}

\paragraph{\method{} is a strong alignment method.}
In \Cref{table:hh_helpful_main}, we can see that \method{} outperforms all baselines. Notably, \method{} consistently shows the highest performance on all three benchmarks while not introducing additional parameters compared to DPO, which is trained in an offline setting. In addition, \method{} also shows significant strength compared to all of the baselines utilizing separate evaluators for off-policy and on-policy learning~\cite{liu2023statistical, gulcehre2023reinforced, dong2023raft, ziegler2020finetuning}. Qualitative examples can be found in \Cref{sec:qualitative_examples}.

\subsection{Analysis of Judge Models}

We conduct an ablation study with HH-Helpful to verify two hypotheses about \method{}: 1) JSFT improves JM's judgment ability, and 2) JM is more compatible with direct preference optimization than RL. To this end, we train three JMs with different strategies: 1) training solely on judgment task induced from \texttt{base} split, 2) JSFT on \texttt{base} split, and 3) JSFT on \texttt{base} split with additional SFT data from \texttt{online} split. We report the performance of each JM as a policy and a judge. We further examine the resulting models on RLHF regarding the predicted likelihood of label tokens on JM as rewards similar to \citet{pmlr-v162-ethayarajh22a}. We also investigate how different sampling strategies in self-training influence the final performance.

\paragraph{JSFT improves JM's judgment ability.}
\Cref{table:hh_helpful_jsft} shows the results regarding the first hypothesis. We can see that JM has only a marginal difference in prediction accuracy on the test split of HH-Helpful compared to RM when it is not trained by JSFT. From this observation, we confirm that the transition effects of imitation learning to judgment task occur when the judgment task is trained with canonical SFT tasks. On the other hand, including the \texttt{online} split for JSFT significantly improves the performance as a policy, but performance as a judge is slightly decreased compared to the JM that is only trained on the \texttt{base} split. We conjecture that the distribution gap between \texttt{online} split and \texttt{base} split influences the transition effects, as \texttt{online} split is obtained from a language model already aligned with human preferences.

\begin{table}[t]
    \centering
    \adjustbox{max width=\columnwidth}{%
    \begin{tabular}{l|c|cc}
        \toprule
        \textbf{Type} & \makecell{\textbf{JSFT} \\ \small{(+ $\mathcal{D}_\text{SFT}$)}} & \makecell{\textbf{Judge} \\ \small{(\% Accuracy)}} & \makecell{\textbf{Policy} \\ \small{(\% Win)}} \\
        \midrule
        RM & \xmark & 66.11 & \xmark \\
        \midrule
        \multirow{3}{*}{JM} & \xmark & 66.49 & \xmark \\
        & \multicolumn{1}{l|}{+ \texttt{base}} &\textbf{68.32} & 5.78 \\
        & + \texttt{base} / \texttt{online} & 67.84 & \textbf{20.26} \\
        \bottomrule
    \end{tabular}}
    \caption{Prediction accuracy on test split of HH-Helpful and winning rate on AlpacaEval using JM as a judge or a policy. JSFT improves not only the performance as a policy but also the performance as a judge of JM.}
    \label{table:hh_helpful_jsft}
    \vspace{-3mm}
\end{table}

\paragraph{On-policy learning with preference orders is the best strategy for JM.}
In \Cref{table:hh_helpful_rlhf}, we observe that RLHF with JM as an evaluator results in a lower performance compared to conventional RM producing scalar rewards, which implies that the judge token likelihood obtained from a judgment could not be regarded as a pointwise preference score on the Bradley-Terry model~\cite{bradley1952rank}. \Cref{table:hh_helpful_dpo} shows that on-policy learning outperforms offline and off-policy learning approaches using a single JM on self-training. Therefore, we can conclude that on-policy learning with preference orders, such as DPO objective, is the best strategy for JM in order to leverage JM's superior performance on judgments compared to the RM.

\begin{table}[ht]
    \centering
    \adjustbox{max width=\columnwidth}{%
    \begin{tabular}{c|c|c|c}
        \toprule
        \textbf{Policy} & \textbf{Evaluator} & \makecell{\textbf{JSFT} \\ \small{(+ $\mathcal{D}_\text{SFT}$)}} & \makecell{\textbf{AlpacaEval} \\ \small{(\% Win)}} \\
        \midrule
        \multirow{4}{*}{SFT} & RM & \xmark & \textbf{33.46} \\
        \cline{2-4}
        & \multirow{3}{*}{JM} & \xmark & 26.18 \\
        & & \multicolumn{1}{l|}{+ \texttt{base}} & 29.63 \\
        & & + \texttt{base} / \texttt{online} & 28.46 \\
        \bottomrule
    \end{tabular}}
    \caption{Evaluation results of models trained with different evaluator types for RLHF. With JMs, we regard the likelihood of judge token comparing with chosen response $y_w$ as a reward\footnotemark. JM's token likelihoods are not appropriate for the pointwise reward function in RLHF.}
    \label{table:hh_helpful_rlhf}
    \vspace{-3mm}
\end{table}
\footnotetext{We found that using the \textit{blank} response fails on the assessment of response since the reward is always close to 1.}

\begin{table}[t]
    \centering
    \adjustbox{max width=\columnwidth}{%
    \begin{tabular}{l|ccc}
        \toprule
        \textbf{Type} & \makecell{\textbf{AlpacaEval}\\ \small{(\% Win)}} & \makecell{\textbf{VicunaEval}\\ \small{(\% Win)}} & \makecell{\textbf{MT-Bench} \\\small{(Score)}} \\
        \midrule
        Offline & 28.57 & 58.75 & 4.68 \\
        Off-Policy & 32.03 & 64.38 & 4.59 \\
        On-Policy & \textbf{44.88} & \textbf{76.25} & \textbf{4.80} \\
        \bottomrule
    \end{tabular}}
    \caption{Effect of learning strategy on self-training. On-policy learning yields the best performance.}
    \label{table:hh_helpful_dpo}
    \vspace{-3mm}
\end{table}

\subsection{Effects of Principle and Rationale}
In this subsection, we compare JMs trained by three different JSFT strategies to verify the effect of principle-aware judgment and rationale through UltraFeedback~\cite{cui2023ultrafeedback}. The three different strategies include 1) judgment derived from overall scores across the principles, 2) principle-aware judgments, and 3) principle-aware judgments with rationales. We check the performance of each JM as a policy and a judge. We also examine how principle-aware judgment and rationale affect the performance as a judge after a self-training stage. We further report the statistics of selected responses on self-rejection according to the sampled number of responses. Additionally, we investigate the feasibility of iterative training, which regards the JM after the self-training as the initial policy and conducts several iterations of self-training.

\begin{table}[h]
    \centering
    \adjustbox{max width=\columnwidth}{%
    \begin{tabular}{l|cc|cc}
        \toprule
        \textbf{Type} & \textbf{P} & \textbf{R} & \makecell{\textbf{Judge} \\ \small{(\% Accuracy)}} & \makecell{\textbf{Policy} \\ \small{(\% Win)}} \\
        \midrule
        RM & \xmark & \xmark & 79.9 & \xmark \\
        \midrule
        JM & \xmark & \xmark & 80.5 & \textbf{67.4} \\
        JM-P & \cmark & \xmark & 81.6 & 59.3 \\
        JM-PR & \cmark & \cmark & \textbf{84.0} & 65.7 \\
        \bottomrule
    \end{tabular}}
    \caption{Performance as a judge or a policy by test split of UltraFeedback and AlpacaEval according to the usage of principle (\textbf{P}) and rationale (\textbf{R}) for pairwise judgment task. Principle-aware judgment with rationale (JM-PR) boosts the performance as a judge while slightly sacrificing the ability as a policy.}
    \label{table:ultrafeedback_principle_rationale}
    \vspace{-3mm}
\end{table}

\begin{figure*}[ht]
  \centering
  \includegraphics[width=\textwidth]{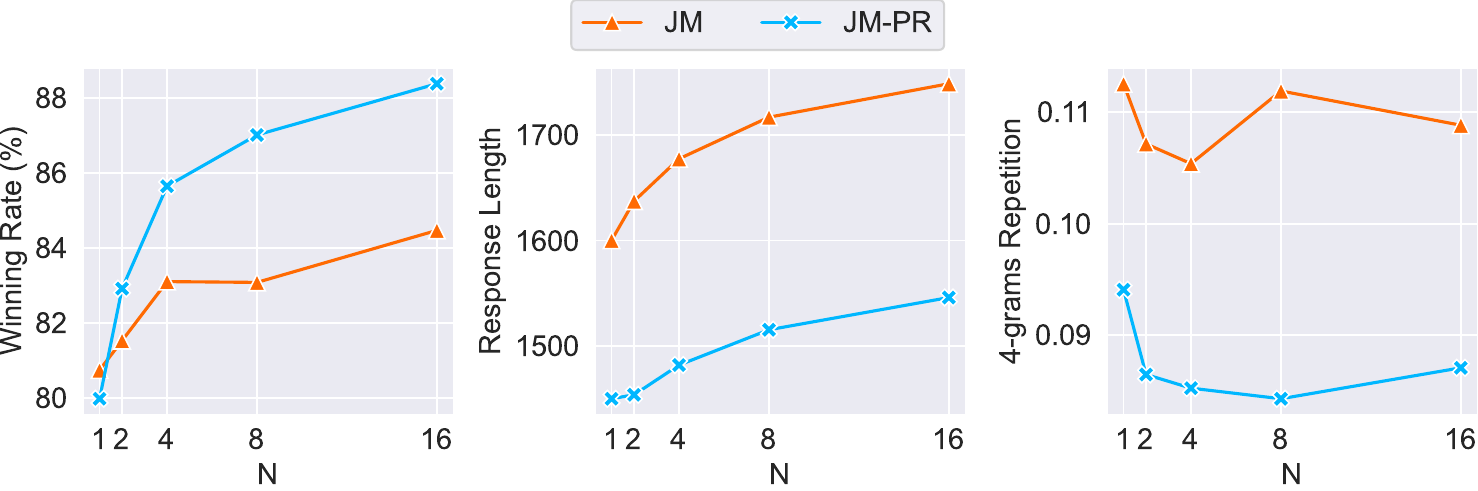}
  \caption{Winning rate, average response length, and 4-gram repetition on AlpacaEval according to the number of sampling ($N$) for self-rejection on JM and JM-PR after self-training. Even though LLM-as-a-judge tends to favor verbose responses~\cite{zheng2023judging}, JM-PR reliably improves the winning rate as $N$ increases, with smoother increments of response lengths and lower repetitions compared to JM.}
  \label{fig:tournament_sampling}
\end{figure*}

\paragraph{Involving principles and rationale improves JM's performance as a judge but not as a policy.}
\Cref{table:ultrafeedback_principle_rationale} shows that JM's performance as a judge increases when the JM is trained for principle-aware judgment (JM-P), and it increases more when the rationale is also used for training (JM-PR). This confirms that the pairwise judgment task can be treated as instruction, inheriting the benefits of instruction-tuning. However, we observe that degeneration in JM's performance as a policy occurs when JM is trained solely on principle-aware judgment but recovered when JM is trained with rationale. We presume this trade-off comes from the bias in task distribution caused by increased pairwise judgment task instances for training principle-aware judgment. However, we speculate that rationale can mitigate performance degradation caused by response distribution mismatch between the pairwise judgment task and other tasks.

\begin{figure}[ht]
  \centering
  \includegraphics[width=\columnwidth]{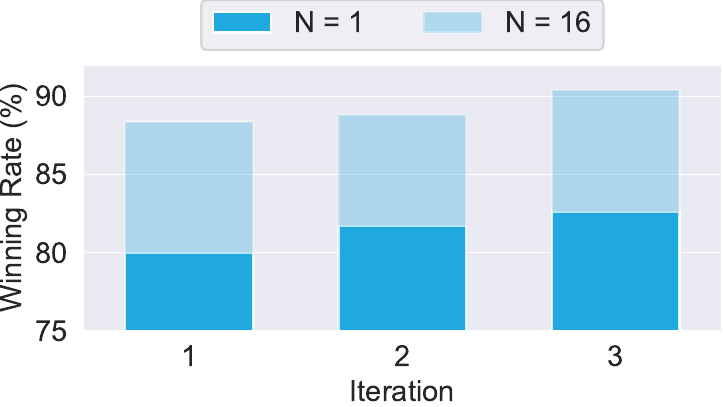}
  \caption{Result of iterative self-training on AlpacaEval using JM-PR. Performance as a policy increases as iterations proceed without losing the capacity as a judge for applying self-rejection.}
  \label{fig:iterative_training}
\end{figure}

\paragraph{JM-PR excels in self-improvement.}
In \Cref{table:ultrafeedback_main}, we can observe that JM-PR achieves comparable performance on benchmarks even though JM-PR shows the degradation as an initial policy model in \Cref{table:ultrafeedback_principle_rationale}. When self-rejection is applied for JM-PR, the winning rate on AlpacaEval is reliably improved up to 8.4\% with shorter response lengths and fewer repetitions compared to JM as \Cref{fig:tournament_sampling} although LLM-as-a-judge is likely to prefer longer responses~\cite{zheng2023judging}. We further experiment with an iterative training scheme that regards the JM-PR after self-training as an initial policy. We observe that the performance as a policy improves while the capability as a judge is maintained, as shown in \Cref{fig:iterative_training}. We find that the increase in performance as a policy diminishes with more iterations, but the performance degeneration as an initial policy can be overcome with iterative training.

\begin{table}[ht]
    \centering
    \adjustbox{max width=\columnwidth}{%
    \begin{tabular}{lcc}
        \toprule
        \textbf{Method} & \makecell{\textbf{AlpacaEval}\\ \small{(\% Win)}} & \makecell{\textbf{MT-Bench} \\\small{(Score)}} \\
        \midrule
        SFT & 68.99 & 5.08 \\
        \midrule
        DPO & \underline{80.25} & 5.74 \\
        RSO & 77.67 & 5.69 \\
        ReST & 72.95 & 5.47 \\
        RAFT & 71.08 & 5.32 \\
        RLHF & 71.68 & 5.40 \\
        \midrule
        \method{} \small{(JM)} & \textbf{80.75} & \underline{6.00} \\
        \textit{+ \small{Self-Rejection ($N$ = 16)}} & 84.47 & 6.08 \\
        \midrule
        \method{} \small{(JM-PR)} & 79.98 & \textbf{6.12} \\
        \textit{+ \small{Self-Rejection} ($N$ = 16)} & 88.39 & 6.14 \\
        \bottomrule
    \end{tabular}}
    \caption{Evaluation results of models trained on UltraFeedback by AlpacaEval and MT-Bench. The best result and second best result without a rejection sampling are represented as bold and underline. Note that applying \textit{self-rejection} on resulted models from \method{} does not require a separate evaluator.}
    \label{table:ultrafeedback_main}
    \vspace{-3mm}
\end{table}

\section{Related Work}

\paragraph{Learning from Preference Scores}
There are several approaches utilizing an RM for alignment learning. RLHF~\cite{ziegler2020finetuning} utilizes an RM for on-policy reinforcement learning. RRHF~\cite{yuan2023rrhf} maximizes the margin of log-likelihood by the rank of responses determined by the score from RM and human annotators. RAFT~\cite{dong2023raft} and ReST~\cite{gulcehre2023reinforced} apply rejection sampling on sampled responses through the RM to perform self-imitation learning. SALMON~\cite{sun2023salmon} trains LLMs to generate scores for responses through principle-driven synthetic preference data utilizing the SFT model. However, all these approaches require a separate RM for the alignment procedure.

\paragraph{Optimizing on Preference Orders}
From preference orders in the static dataset, DPO~\cite{rafailov2023direct} optimizes LLMs by implicit rewards without a separated RM. IPO~\cite{azar2023general} proposes a modified objective using an unbounded preference mapping function to mitigate overfitting on deterministic preferences in the dataset. PCO~\cite{xu2023some} utilizes cringe loss for optimization, which considers the token-level likelihood of rejected samples as contrastive training. SPIN~\cite{chen2024self} performs iterative training considering the distribution gap between SFT datasets and the model's responses as preference orders. Self-Rewarding Language Models~\cite{yuan2024self} trains LLMs to generate scores for a given response by chain-of-thought reasoning to construct preference datasets by self-generated responses. All these approaches differ from our work in that they do not perform on-policy learning.

\paragraph{Generative Pairwise Evaluator}
The generative pairwise evaluator, which we refer to as JM, has been utilized in previous approaches to alignment learning. ILF~\cite{scheurer2023training} selects the response that reflects human-requested feedback through JM. SLiC-HF~\cite{zhao2023slic} constructs a static preference dataset with responses obtained from the SFT model ordered by JM. RSO~\cite{liu2023statistical} approximates the optimal policy of the RLHF objective with rejection sampling through JM's judge token likelihood. OAIF~\cite{guo2024direct} adopts pre-aligned LLMs to perform the pairwise judgment task for on-policy learning, not by fine-tuning to JM. All of these approaches focus on utilizing a separate evaluator and do not address self-improvement.
\section{Discussion}
\paragraph{Concurrent Works}
Recent works such as Self-Rewarding Language Models~\cite{yuan2024self} and OAIF~\cite{guo2024direct} have intersections with \method{} in aspects of self-training and on-policy learning for pairwise preferences. Self-Rewarding Language Models trains the initial policy to act as a judge, but evaluations are performed through chain-of-thought reasoning, which requires a significant computational cost for evaluation. Therefore, off-policy learning with an iterative training scheme was adopted, which can yield inferior results compared to on-policy learning, as shown in \Cref{table:hh_helpful_dpo}. OAIF similarly utilizes a large language model as a judge for on-policy learning but employs an already aligned large language model as the judge. This necessitates a superior aligned language model and incurs additional computational resources compared to this work, which performs self-improvement using a single model. \Cref{table:concurrent_works} illustrates the difference between these approaches compared to \method{}.

\begin{table}[t]
    \centering
    \adjustbox{max width=\columnwidth}{%
    \begin{tabular}{c|ccc}
        \toprule
        \textbf{Method} & \textbf{Self-Training} & \textbf{On-Policy} \\
        \midrule
        \makecell{Self-Rewarding LMs \\ \small{\cite{yuan2024self}}} & \cmark & \xmark \\
        \midrule
        \makecell{OAIF \\ \small{\cite{guo2024direct}}} & \xmark & \cmark \\
        \midrule
        \makecell{\textbf{\method{}} \\ \small{(Ours)}} & \cmark & \cmark \\
        \bottomrule
    \end{tabular}}
    \caption{Comparison between \method{} and recent concurrent works on preference alignment learning. Unlike the others, \method{} achieves both self-training without separated evaluators and on-policy learning.}
    \label{table:concurrent_works}
\end{table}

\paragraph{Effects of Parameter Size on JSFT}
\citet{gao2023scaling} demonstrates that scaling of parameter size enhances the robustness of RMs in the sense of proxy for the ground truth reward. Similarly, we believe parameter size might significantly affect JMs since they act as a proxy for pairwise human preferences. \Cref{table:hh_helpful_params} shows that as the parameter size increases, the performance of JM increases without JSFT, but the performance gain from JSFT grows correspondingly as parameter size increases. This implies that the transition effects of imitation learning to judgment task positively correlate with parameter size. Therefore, \method{} has the additional advantage of obtaining a strong evaluator from the overparameterization of the policy model, since JSFT directly leads the scaling of the initial policy to the scaling of the evaluator.

\begin{table}[h]
    \centering
    \adjustbox{max width=\columnwidth}{%
    \begin{tabular}{l|c|cc}
        \toprule
        \textbf{Model} & \textbf{JSFT} & \makecell{\textbf{Judge} \\ \small{(\% Accuracy)}} & \makecell{\textbf{Policy} \\ \small{(\% Win)}} \\
        \midrule
        \multirow{2}{*}{Llama-2-7B} & \xmark & 66.11 & \xmark \\
        & \cmark & 67.84 & 20.26 \\
        \midrule
        \multirow{2}{*}{Llama-2-13B} & \xmark & 66.86 & \xmark \\
        & \cmark & \textbf{72.13} & \textbf{24.00} \\
        \bottomrule
    \end{tabular}}
    \caption{Prediction accuracy on test split of HH-Helpful and winning rate on AlpacaEval with different sizes of base model's parameter size for JMs. We use \texttt{base} and \texttt{online} splits of HH-Helpful for JSFT.}
    \label{table:hh_helpful_params}
    \vspace{-3mm}
\end{table}

\paragraph{Policy Model as a Judge for Self-Training}
The reference model is regarded as a judge for the policy model on the self-training by default in \method{}. This corresponds to the assumption of a static environment during a single iteration of the self-training stage. On the other hand, \Cref{fig:iterative_training} shows the policy model's capacity as a judge remained after the self-training stage. This implies that self-training can be conducted even in a dynamic environment by regarding the policy model as a judge whose judgment capability may be improved during the self-training process. However, we find that the policy model as a judge yields slightly lower performances compared to the reference model, as shown in \Cref{table:hh_helpful_judge}. Therefore, we can conclude that the static evaluator has benefits in the performance, while the dynamic evaluator still can be utilized with comparable performance.

\begin{table}[ht]
    \centering
    \adjustbox{max width=\columnwidth}{%
    \begin{tabular}{c|c|cc}
        \toprule
        \textbf{Policy} & \textbf{Evaluator} & \makecell{\textbf{AlpacaEval}\\ \small{(\% Win)}} & \makecell{\textbf{MT-Bench} \\\small{(Score)}} \\
        \midrule
        \multirow{3}{*}{JM} & \xmark & 20.26 & 4.19 \\
        \cline{2-4}
        & $\pi_{\theta}$ & 42.38 & 4.67 \\
        & $\pi_{\text{ref}}$  & \textbf{44.88} & \textbf{4.80} \\
        \bottomrule
    \end{tabular}}
    \caption{Evaluation results of models trained on HH-Helpful with different strategies on a judge for the self-training stage. It shows that slight degeneration happens when utilizing the policy model $\pi_{\theta}$ as a judge compared to the reference model $\pi_{\text{ref}}$ as a judge.}
    \label{table:hh_helpful_judge}
    \vspace{-3mm}
\end{table}
\section{Conclusion}

We propose a parameter-efficient on-policy preference alignment framework, \method{}, introducing Judge-augmented Supervised Fine-tuning (JSFT). A model trained by JSFT can perform feedback to the current policy for improving itself by acting as a judge. This self-training does not require additional training stages and parameters for a reward model during the policy updates. Our resulting model outperforms RLHF, offline, and off-policy baselines in preference benchmarks, demonstrating the advantages of on-policy learning and parameter efficiency of \method{}. Besides, we provide various analyses on the best configurations and efficacy of the proposed JSFT. Specifically, JSFT boosts performance as a judge, and involving comparisons based on principle with rationale about decision leads to further improvement. This enhanced judging capability leads to further self-improvement as a policy at inference time by self-rejection over the model's own responses.

\section*{Limitations}

To achieve parameter-efficient on-policy self-training, \method{} assumes the presence of a human preference dataset of examples for pairwise judgment tasks on JSFT. Therefore, if human preference datasets~\cite{bai2022training} or strong teacher models for constructing AI Feedback datasets~\cite{cui2023ultrafeedback} are not available, \method{} can not be utilized. This means that \method{} has a limitation compared to self-alignment approaches that can construct a preference dataset when there is no preference dataset at all~\cite{bai2022constitutional, sun2023salmon, yuan2024self}. Additionally, our experiments do not focus on safety. Thus, using  \method{} without reviewing safety guards may lead to potentially inappropriate responses.
\section*{Acknowledgements}

This work was partly supported by an IITP grant funded by the Korean Government
(MSIT) (No. RS-2020-II201361, Artificial Intelligence Graduate School Program (Yonsei
University)) and the National Research Foundation of Korea(NRF) grant funded by the Korea government(MSIT) (No. RS-2024-00354218). The authors thank the members of NAVER Cloud and Yonsei University for their constructive comments. We are also grateful to
Seungju Han for valuable discussions and helpful
feedback on earlier drafts of this paper.

\bibliography{anthology,custom}
\bibliographystyle{acl_natbib}

\clearpage

\appendix
\onecolumn
\section{Implementation Details}
\label{sec:implementation_details}

\subsection{Pre-processing of Datasets}

\begin{table}[h]
    \centering
    \adjustbox{max width=\columnwidth}{%
    \begin{tabular}{l|cc}
        \toprule
        \textbf{Target} & \textbf{HH-Helpful} & \textbf{UltraFeedback} \\
        \midrule
        SFT & 65842 & 57569 \\
        \midrule
        Preference & 43835 & 57266 \\
        \small(+ Principle) & \xmark & 202887 \\
        \bottomrule
    \end{tabular}}
    \caption{The number of training examples in each dataset according to the target datasets.}
    \label{table:training_examples}
\end{table}

For the HH-Helpful\footnote{\href{https://huggingface.co/datasets/Anthropic/hh-rlhf}{huggingface.co/datasets/Anthropic/hh-rlhf}, MIT License, Copyright (c) 2022 Anthropic}~\cite{bai2022training}, we parse only the content of each turn through the role header in the dataset itself. During this process, if a role header exists redundantly (e.g., \textit{Human: Assistant: [content]}), we remove all subsequent headers. We perform the roll-out procedures from the last assistant turn in each dialogue. For the UltraFeedback\footnote{\href{https://huggingface.co/datasets/openbmb/UltraFeedback}{huggingface.co/datasets/openbmb/UltraFeedback}, MIT License, Copyright (c) 2023 THUNLP}~\cite{cui2023ultrafeedback}, we use the mean score across principles as the overall rank of the responses. We choose the longer response as a better response when we have a tie~\cite{hosking2023human}. We randomly sample one response as a rejected response $y_l$ that is inferior in rank or score on each principle for principle-aware judgment. When a rationale is utilized in training, we remove responses that include a comparative explanation against other responses. \Cref{table:training_examples} shows the number of resulted training examples on each dataset after the described pre-processing.

\subsection{Hyperparameters}

\begin{table}[h]
    \centering
    \adjustbox{max width=\columnwidth}{%
    \begin{tabular}{l|cc}
        \toprule
        \textbf{Hyperparameters} & \textbf{Initial} & \textbf{Feedback} \\
        \midrule
        Epoch & 1 & 3\\
        Batch Size & 128 & 64 \\
        Learning Rate & 2e-5 & 5e-6 \\
        LR Scheduler & cosine & constant \\
        Warm-up Ratio & 0.03 & 0.1 \\
        Temperature & \xmark & 1.0 \\
        Top-p & \xmark & 0.9 \\
        Max New Tokens & \xmark & 768 \\
        \midrule
        Optimizer & \multicolumn{2}{c}{AdamW} \\
        $(\beta_1, \beta_2)$ & \multicolumn{2}{c}{(0.9, 0.999)} \\
        Gradient Clipping & \multicolumn{2}{c}{1.0} \\
        Max Sequence Length & \multicolumn{2}{c}{2048} \\
        \bottomrule
    \end{tabular}
    \hspace{1cm}
    \begin{tabular}{l|c}
        \toprule
        \multicolumn{2}{c}{\textbf{LoRA}} \\
        \midrule
        (r, $\alpha$) & (8, 16) \\
        Dropout & 0.1 \\
        \midrule
        \multicolumn{2}{c}{\textbf{RLHF}} \\
        \midrule
        Mini-batch Size & 32 \\
        Inner Epochs & 1 \\
        KL Scheduler & (0.2, 6.0) \\
        \midrule
        \multicolumn{2}{c}{\textbf{ReST}} \\
        \midrule
        $\tau$ & [0.7, 0.8, 0.9] \\
        \midrule
        \multicolumn{2}{c}{\textbf{DPO, RSO, \method{}}} \\
        \midrule
        $\beta$ & 0.1 \\
        \bottomrule
    \end{tabular}}
    \caption{Hyperparameters of the experiments. \textit{Initial} refers to the value of hyperparameters for SFT, RM, and JM. \textit{Feedback} refers the value of hyperparameters for baselines and \method{}.}
    \label{table:hyperparameters}
\end{table}

We apply ReST~\cite{gulcehre2023reinforced} with $G=1, I=3$, regarding a single step of Improve step as one epoch of training and $\tau$ as a quantile threshold on reward distribution. We sample 8 responses per prompt for RAFT~\cite{dong2023raft} and RSO~\cite{liu2023statistical}. We randomly choose one of the sampled responses as a baseline response for JMs and accepted a maximum of 1 response per prompt for RSO. We do not conduct a hyperparameter search. Table \ref{table:hyperparameters} shows hyperparameters used in the experiments.

\clearpage

\subsection{Training Details}

We perform full fine-tuning to obtain the initial policy and evaluators, RM or JM initializing from Llama-2-7B\footnote{\href{https://huggingface.co/meta-llama/Llama-2-7b}{huggingface.co/meta-llama/Llama-2-7b}, LLAMA 2 Community License, Copyright (c) 2023 Meta Platforms}~\cite{touvron2023llama2}. We apply LoRA~\cite{hu2021lora} for computational efficiency of fine-tuning the baselines and \method{}. We calculate language modeling loss on responses of assistant for SFT and sequences after the judgment template $\mathcal{C}$ for the judgment task. We train JM with both position-swapped comparisons on the judgment template for given triplet $(x, y_w, y_l)$ to avoid position bias~\cite{zheng2023judging}. We use the prompt $x$ used on the pairwise preference comparison for roll-out procedures. Every experiment is conducted with a maximum of 8 NVIDIA A100 80GB VRAM GPUs for up to 90 hours. The baselines are implemented based on the Transformers\footnote{\href{https://github.com/huggingface/transformers}{github.com/huggingface/transformers}}, PEFT\footnote{\href{https://github.com/huggingface/peft}{github.com/huggingface/peft}} and TRL\footnote{\href{https://github.com/huggingface/trl}{github.com/huggingface/trl}} library.

\subsection{Evaluation Details}

We use the checkpoints from the end of training for evaluation without any model selections on every method. Since VicunaEval~\cite{vicuna2023} and MT-Bench~\cite{zheng2023judging} contain question types that are forced to use greedy decoding in the pre-defined configuration for each benchmark, we only apply the self-rejection on questions that do not require the greedy decoding on evaluation.

\subsection{Prompt Templates}

\begin{table*}[ht]
\centering
\small
\noindent\fbox{
\begin{varwidth}{\dimexpr\linewidth-2\fboxsep-2\fboxrule}
\tt
<|system|>\\
\{system message\}\\
<|user|> \\
\{prompt at turn 1\}\\
<|assistant|>\\
\{response at turn 1\}[EOS]\\
<|user|>\\
\{prompt at turn 2\}\\
<|assistant|>\\
\{response at turn 2\}[EOS]\\
$\cdots$
\end{varwidth}
}
\caption{The dialogue template for the experiments.}
\label{table:dialogue_template}
\end{table*}
\begin{table*}[ht]
\centering
\small
\noindent\fbox{
\begin{minipage}{\dimexpr\linewidth-2\fboxsep-2\fboxrule}
\textbf{HH-Helpful}\\
\\
\texttt{Below are a series of dialogues between a human user and an AI assistant. The AI tries to be helpful, harmless, and honest.}\\
\\
\textbf{UltraFeedback} \\
\\
\texttt{Below are a series of dialogues between a human user and an AI
assistant. The AI tries to be helpful, honest, truthful and follows the
instructions from human.}
\end{minipage}
}
\caption{The default system messages for each dataset.}
\label{table:default_system}
\end{table*}

\Cref{table:dialogue_template} shows the dialogue template used for the experiments and \Cref{table:default_system} shows default system messages used for each dataset. We use the default system message for both training and inference on HH-Helpful. In the case of UltraFeedback, we use a custom system prompt for the chosen response in the dataset when constructing training examples. We use the default system prompt for training on the judgment task and inference. For the judgment template $\mathcal{C}$, we use the template shown in \Cref{table:hh_judgment_template} for HH-Helpful and the template shown in \Cref{table:uf_judgment_template} for UltraFeedback. For the principle-aware judgments on UltraFeedback, we modify the guidance about principles as a system prompt as shown in \Cref{table:principle_system}~ \cite{cui2023ultrafeedback}. We use the principle-aware judgment template $\mathcal{C}_p$ shown in \Cref{table:uf_judgment_p_template} and slightly modify this when rationale $\mathcal{R}$ is included in the training, as shown in \Cref{table:uf_judgment_pr_template}.

\clearpage

\begin{table*}[t]
\centering
\small
\noindent\fbox{
\begin{minipage}{\dimexpr\linewidth-2\fboxsep-2\fboxrule}
\texttt{<|system|>}\\
\texttt{\{default system message\}}\\
\texttt{<|user|>}\\
\texttt{Which is the more helpful, harmless, and honest assistant from a given context?}\\
\\
\texttt{Context:}\\
\texttt{-------}\\
\texttt{User: \{prompt at turn 1\}}\\
\\
\texttt{Assistant: \{responses at turn 1\}}\\
\\
\texttt{$\cdots$}\\
\\
\texttt{User: \{prompt at turn N\}}\\
\texttt{-------}\\
\\
\texttt{Assistant A: \{responses A at turn N\}}\\
\texttt{Assistant B: \{responses B at turn N\}}\\
\\
\texttt{Please choose either A or B.}\\
\texttt{<|assistant|>}\\
\texttt{Sure! The option which is more helpful, harmless, and honest would be Assistant } \textbf{\{\texttt{A or B}\}}
\end{minipage}
}
\caption{The judgment template $\mathcal{C}$ for HH-Helpful. The target sequence for the training judgment task is in bold. The default system message is shown in \Cref{table:default_system}.}
\label{table:hh_judgment_template}
\end{table*}
\begin{table*}[t]
\centering
\small
\noindent\fbox{
\begin{minipage}{\dimexpr\linewidth-2\fboxsep-2\fboxrule}
\texttt{<|system|>}\\
\texttt{\{default system message\}}\\
\texttt{<|user|>}\\
\texttt{Which is the better response to be an assistant who is helpful, honest, truthful and following the given instruction
from user?}\\
\\
\texttt{Instruction:}\\
\texttt{-------}\\
\texttt{\{prompt\}}\\
\texttt{-------}\\
\\
\texttt{Response A:}\\
\texttt{-------}\\
\texttt{\{response A\}}\\
\texttt{-------}\\
\\
\texttt{Response B:}\\
\texttt{-------}\\
\texttt{\{response B\}}\\
\texttt{-------}\\
\\
\texttt{Please choose either A or B.}\\
\texttt{<|assistant|>}\\
\texttt{Sure! The option which is more helpful, honest, truthful and following the given instruction from user would be
Response } \textbf{\{\texttt{A or B}\}}\\
\end{minipage}
}
\caption{The judgment template $\mathcal{C}$ for UltraFeedback. The target sequence for the training judgment task is in bold. The default system message is shown in \Cref{table:default_system}.}
\label{table:uf_judgment_template}
\end{table*}

\clearpage

\begin{table*}[t]
\centering
\small
\noindent\fbox{
\begin{minipage}{\dimexpr\linewidth-2\fboxsep-2\fboxrule}
\textbf{Helpfulness}\\
\\
\texttt{Under the principle of 'helpfulness', the assistant should provide users with accurate, relevant, and up-to-date
information, ensuring that the content is positive, interesting, engaging, educational, and helpful.}\\
\\
\textbf{Honesty}\\
\\
\texttt{Under the principle of 'honesty', the assistant should be honest about whether it knows the answer and express its
uncertainty explicitly. The assistant should confident on questions it knows well and be modest on those it is
unfamiliar with using weakeners such as 'I guess', 'I suppose', 'probably', and 'perhaps' to express uncertainty.}\\
\\
\textbf{Instruction Following}\\
\\
\texttt{Under the principle of 'instruction following', the assistant should align the output with intent of instruction, by
understanding the task goal (intended outcome) and restrictions (text styles, format or designated methods, etc.).}\\
\\
\textbf{Truthfulness}\\
\\
\texttt{Under the principle of 'truthfulness', the assistant should answer truthfully and be faithful to factual knowledge as
well as given contexts, never making up any new facts that aren’t true or cannot be grounded in the instruction.}
\end{minipage}
}
\caption{The principle-aware system messages for UltraFeedback.}
\label{table:principle_system}
\end{table*}
\begin{table*}[t]
\centering
\small
\noindent\fbox{
\begin{minipage}{\dimexpr\linewidth-2\fboxsep-2\fboxrule}
\texttt{<|system|>}\\
\texttt{\{principle-aware system message\}}\\
\texttt{<|user|>}\\
\texttt{Which is the better response for an assistant when following the principle of '\{principle\}' for a given instruction?}\\
\\
\texttt{Instruction:}\\
\texttt{-------}\\
\texttt{\{prompt\}}\\
\texttt{-------}\\
\\
\texttt{Response A:}\\
\texttt{-------}\\
\texttt{\{response A\}}\\
\texttt{-------}\\
\\
\texttt{Response B:}\\
\texttt{-------}\\
\texttt{\{response B\}}\\
\texttt{-------}\\
\\
\texttt{Please choose either A or B according to the principle of '\{principle\}'.}\\
\texttt{<|assistant|>}\\
\texttt{Sure! The option which is better guided by the principle of '\{principle\}' would be Response } \textbf{\{\texttt{A or B}\}}\\
\end{minipage}
}
\caption{The principle-aware judgment template $\mathcal{C}_p$ for UltraFeedback. The target sequence for the training judgment task is in bold. The principle-aware system messages are shown in \Cref{table:principle_system}.}
\label{table:uf_judgment_p_template}
\end{table*}
\begin{table*}[t]
\centering
\small
\noindent\fbox{
\begin{minipage}{\dimexpr\linewidth-2\fboxsep-2\fboxrule}
\texttt{<|system|>}\\
\texttt{\{principle-aware system message\}}\\
\texttt{<|user|>}\\
\texttt{Which is the better response for an assistant when following the principle of '\{principle\}' for a given instruction?}\\
\\
\texttt{Instruction:}\\
\texttt{-------}\\
\texttt{\{prompt\}}\\
\texttt{-------}\\
\\
\texttt{Response A:}\\
\texttt{-------}\\
\texttt{\{response A\}}\\
\texttt{-------}\\
\\
\texttt{Response B:}\\
\texttt{-------}\\
\texttt{\{response B\}}\\
\texttt{-------}\\
\\
\texttt{Please choose either A or B and evaluate each response according to the principle of '\{principle\}'.}\\
\texttt{<|assistant|>}\\
\texttt{Sure! Here is the decision of better response and evaluations of each response, guided by the principle of '\{principle\}'.}\\
\\
\texttt{Decision: Response } \textbf{\{\texttt{A or B}\}}\\
\\
\textbf{\texttt{Evaluation of Response A: \{rationale A\}}}\\
\textbf{\texttt{Evaluation of Response B: \{rationale B\}}}
\end{minipage}
}
\caption{The principle-aware judgment template $\mathcal{C}_p$ when rationale $\mathcal{R}$ is included in UltraFeedback. The target sequence for the training judgment task is in bold.}
\label{table:uf_judgment_pr_template}
\end{table*}

\clearpage

\section{Qualitative Examples}
\label{sec:qualitative_examples}

We compare responses of the resulting models from baseline and \method{} trained on HH-Helpful. We further compare the difference after self-training between JM and JM-PR on UltraFeedback by the results of self-rejection to understand the effects of principle-aware judgment and rationale. In \Cref{table:baseline_comparisons}, we observe that \method{} provides richer information requested by the prompt compared to other baselines. Furthermore, we confirm that JM is likely to select a response that contains unnecessarily verbose sentences as the best response compared to JM-PR, as shown in \Cref{table:pr_tendency}.

\begin{table*}[h]
\centering
\small
\noindent\fbox{
\begin{minipage}{\dimexpr\linewidth-2\fboxsep-2\fboxrule}
\textbf{Prompt}\\
\\
\texttt{what is the name of chris tucker first movie}
\end{minipage}
}
\vspace{0.3cm}\\
\noindent\fbox{
\begin{minipage}{\dimexpr\linewidth-2\fboxsep-2\fboxrule}
\textbf{DPO}\\
\\
\texttt{The name of Chris Tucker’s first movie is “Friday”.  It was released in 1995.  The movie was written by Tucker and Ice Cube, and also stars Ice Cube and Nia Long.}\\
\\
\textbf{RSO}\\
\\
\texttt{Chris Tucker’s first major film role was in the 1995 comedy film “Friday”.  He also had roles in “The Fifth Element” and “Money Train”.  His most famous films include “Rush Hour”, “The Best Man”, and “Rush Hour 2”.}\\
\\
\textbf{ReST}\\
\\
\texttt{The name of Chris Tucker’s first movie was “Friday”.}\\
\\
\textbf{RAFT}\\
\\
\texttt{The name of Chris Tucker’s first movie is “Friday”, which was released in 1995.}\\
\\
\textbf{RLHF}\\
\\
\texttt{His first film was "Friday", released in 1995.  The film was a great success, and helped launch him to fame.  It earned him a Golden Globe nomination for Best Actor in a Musical or Comedy, and received critical acclaim.  It is considered a classic modern comedy movie, and has become a cult film.  A sequel was released in 2007, and a third film in the franchise is planned for 2017.}\\
\\
\textbf{\method{}}\\
\\
\texttt{The first movie that Chris Tucker starred in was the 1995 comedy film “Friday”, directed by F. Gary Gray.  \textbf{The film is a classic urban comedy, centering around the adventures of two friends, played by Tucker and Ice Cube, as they spend the day hanging out in their Los Angeles neighborhood.}  The film was a breakout hit for Tucker, establishing him as a prominent comedic actor, and was followed by a successful sequel in 1997.  Tucker then went on to star in several major Hollywood films, including the hit comedy films “Boomerang” and “Tommy” as well as the drama “Rush Hour” and its sequels.  He has also had a successful career as a stand-up comedian, performing sold-out shows worldwide.
}
\end{minipage}
}
\caption{The response examples of resulting models from each method on AlpacaEval.}
\label{table:baseline_comparisons}
\end{table*}
\clearpage
\begin{table*}[t]
\centering
\small
\noindent\fbox{
\begin{minipage}{\dimexpr\linewidth-2\fboxsep-2\fboxrule}
\textbf{Prompt}\\
\\
\texttt{Take a look at the contract and extract the parties of the agreement from it.}\\
\\
\texttt{As of the 13th day of December 2021, this Music Recording Agreement ("Agreement") is made between Good Kid, a Toronto-based musical group ("Artist"), and Universal Music Group, a record label with license number 545345 ("Record Label"). The Artist and the Recording Label may be referred to in this Agreement individually as "Parties" and collectively as "Parties." Work under this Agreement will begin on March 15, 2022, and will last for the duration of the Agreement.}
\end{minipage}
}
\vspace{0.3cm}\\
\noindent\fbox{
\begin{minipage}{\dimexpr\linewidth-2\fboxsep-2\fboxrule}
\textbf{JM ($\mathbf{N}$ = 1)}\\
\\
\texttt{The parties of the Music Recording Agreement are:}\\
\\
\texttt{1. Good Kid, a Toronto-based musical group ("Artist")}\\
\texttt{2. Universal Music Group, a record label with license number 545345 ("Record Label")}\\
\\
\texttt{Both parties are referred to individually as "Parties" in the Agreement and collectively as "Parties." This means that the Artist and the Record Label are the two main participants in the agreement.}\\
\\
\textbf{JM-PR ($\mathbf{N}$ = 1)}\\
\\
\texttt{The parties involved in the Music Recording Agreement are:}\\
\\
\texttt{* Good Kid, a Toronto-based musical group, represented by the artist.}\\
\texttt{* Universal Music Group, a record label with license number 545345, represented by the record label.)}\\
\\
\texttt{This agreement is effective as of the 13th day of December 2021 and will last until the expiration of the agreement. The work under this agreement will begin on March 15, 2022, and will continue for the duration of the agreement.}
\end{minipage}
}
\vspace{0.3cm}\\
\noindent\fbox{
\begin{minipage}{\dimexpr\linewidth-2\fboxsep-2\fboxrule}
\textbf{JM ($\mathbf{N}$ = 16)}\\
\\
\texttt{Based on the provided Music Recording Agreement, the parties involved in the agreement are:}\\
\\
\textbf{\texttt{1. Good Kid, a Toronto-based musical group (Artist).}}\\
\textbf{\texttt{2. Universal Music Group, a record label with license number 545345 (Record Label).}}\\
\\
\texttt{The Agreement refers to the Artist and the Record Label as "Parties" and collectively as "Parties" throughout the document. \textbf{Therefore, the parties to the Music Recording Agreement are:}}\\
\\
\textbf{\texttt{* Good Kid, a Toronto-based musical group (Artist)}}\\
\textbf{\texttt{* Universal Music Group, a record label with license number 545345 (Record Label)}}\\
\\
\texttt{The Agreement outlines the terms and conditions of the recording agreement between the Artist and the Record Label, including the commencement date, the duration of the agreement, and the scope of work.}\\
\\
\textbf{JM-PR ($\mathbf{N}$ = 16)}\\
\\
\texttt{Based on the provided contract, the parties involved in the Music Recording Agreement are:}\\
\\
\texttt{1. Good Kid, a Toronto-based musical group ("Artist")}\\
\texttt{2. Universal Music Group, a record label with license number 545345 ("Record Label")}\\
\\
\texttt{These two parties are involved in the agreement and are referred to individually as "Parties" and collectively as "Parties" throughout the document.}
\end{minipage}
}
\caption{The response examples of JM and JM-PR by applying self-rejection on AlpacaEval.}
\label{table:pr_tendency}
\end{table*}

\end{document}